\documentclass{article}

\PassOptionsToPackage{numbers, compress}{natbib}

\usepackage[preprint]{neurips_2025}

\usepackage[utf8]{inputenc} 
\usepackage[T1]{fontenc}    
\usepackage{hyperref}       
\usepackage{url}            
\usepackage{booktabs}       
\usepackage{amsfonts}       
\usepackage{nicefrac}       
\usepackage{microtype}      
\usepackage{xcolor}         

\usepackage{hhline}
\usepackage{subcaption}
\usepackage{pifont}
\usepackage{bm}

\usepackage{multirow}
\usepackage{floatrow}\newfloatcommand{capbtabbox}{table}[][\FBwidth]
\usepackage{pifont}
\usepackage{color}
\usepackage{anyfontsize}
\usepackage[11pt]{moresize}
\usepackage[export]{adjustbox}
\usepackage{amssymb}
\usepackage{mathrsfs}
\usepackage{amsmath}

\usepackage{algorithm}
\usepackage{algpseudocode}  
\usepackage{listings}
\usepackage{multirow}

\title{UHD Image Dehazing via anDehazeFormer with Atmospheric-aware KV Cache}

\author{
    Pu Wang \\
    Shandong University\\
   \And
   Pengwen Dai \\
   Sun Yat-sen University  \\
   \AND
   Chen Wu \\
   USTC \\
   \And
   Yeying Jin \\
   Tencent \\
   \And
   Dianjie Lu \\
   SDNU\\
   \And
   Guijuan Zhang \\
   SDNU \\
   \And
   Youshan Zhang \\
   Yeshiva University\\
   \And
   Zhuoran Zheng\thanks{Corresponding Author, \texttt{zhengzhr@mail.sysu.edu.cn}} \\
   Sun Yat-sen University \\
}

\begin{document}

\bibliographystyle{plain}
\maketitle

\begin{abstract}
In this paper, we propose an efficient visual transformer framework for ultra-high-definition (UHD) image dehazing that addresses the key challenges of slow training speed and high memory consumption for existing methods. Our approach introduces two key innovations: 1) an \textbf{a}daptive \textbf{n}ormalization mechanism inspired by the nGPT architecture that enables ultra-fast and stable training with a network with a restricted range of parameter expressions; and 2) we devise an atmospheric scattering-aware KV caching mechanism that dynamically optimizes feature preservation based on the physical haze formation model. The proposed architecture improves the training convergence speed by \textbf{5 $\times$} while reducing memory overhead, enabling real-time processing of 50 high-resolution images per second on an RTX4090 GPU. Experimental results show that our approach maintains state-of-the-art dehazing quality while significantly improving computational efficiency for 4K/8K image restoration tasks. Furthermore, we provide a new dehazing image interpretable method with the help of an integrated gradient attribution map. Our code can be found here: \url{https://anonymous.4open.science/r/anDehazeFormer-632E/README.md}.

\end{abstract}

\section{Introduction}%
The exponential growth of ultra-high-definition (UHD) imaging techniques, ranging from 4K medical endoscopes to 8K satellite sensors, has significantly transformed visual data acquisition across various industries. While these techniques can capture unprecedented levels of detail, their operational environments frequently introduce atmospheric distortions~\cite{sahu2022trends, yamashita2016ultra} in natural environments.  Traditional dehazing methods encounter a critical efficiency paradox: physics-based approaches \cite{shrivastava2016single,muniyappan2013novel} struggle to address the complexities associated with real-world scattering phenomena. In contrast, deep learning methods \cite{choudhary2020image,wang2024correlation,kim2018real} require substantial computational resources for UHD processing, typically necessitating 80GB of GPU memory for handling a 4K image.

Current solutions try to strike a balance between quality and efficiency by processing images in patches \cite{darabi2012image} or by downsampling the resolution \cite{mac2012patch}. However, these approaches come with two major drawbacks: 1) Localized processing messes up the overall atmospheric consistency, leading to noticeable seams in sky areas; 2) The major challenge is to create a UHD dehazing algorithm that maintains the perception of the spatial environment while preserving the regression accuracy at different scale resolutions on low computational resources.

\begin{figure}[t]
    \centering
    \includegraphics[width=0.975\textwidth]{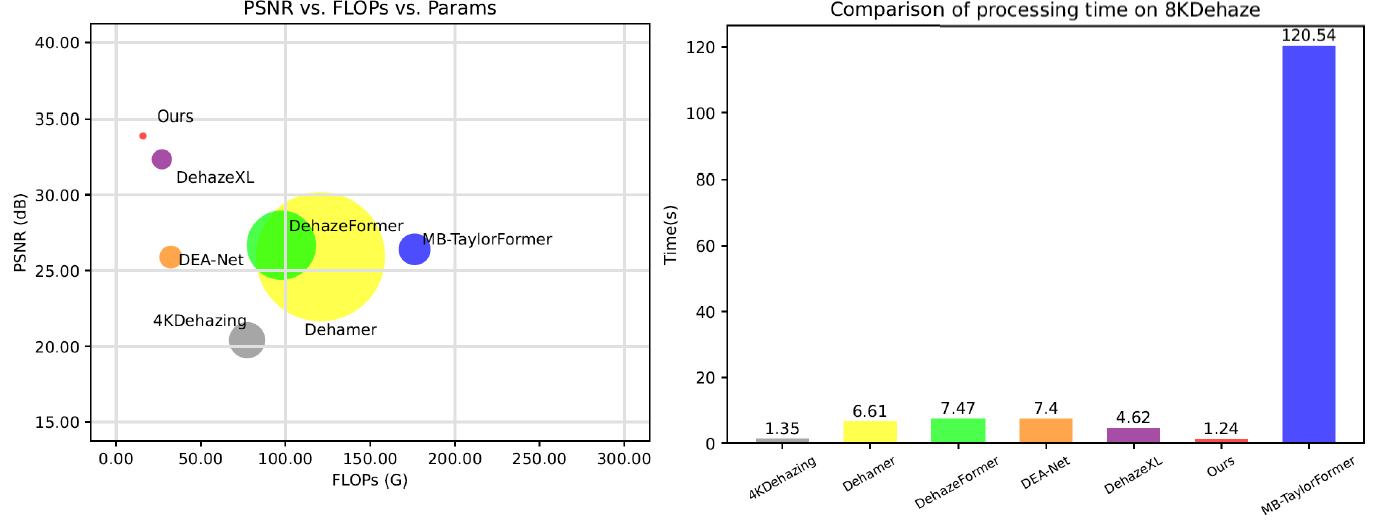}
    \caption{Improvement of our model over the SOTA approaches in 8KDehaze~\cite{chen2025tokenize0}. The circle size is proportional to the number of model parameters.}
    \label{fig:out}
    \vspace{-4mm}
\end{figure}

To address these challenges, we introduce anDehazeFormer, an efficient vision Transformer method that revolutionizes UHD image dehazing by integrating adaptive normalization and physics-aware cache mechanisms.  
Specifically, our method leverages two key innovations to overcome the limitations of conventional methods.
i) \textbf{Adaptive Normalization} for stable and rapid training:  We are inspired by the idea of normalizing all parameters of a neural network proposed by nGPT, but the difference is that we are adaptive on each network block. As shown in Fig.~\ref{fig:out}, by leveraging the physical parameters learned by the atmosphere-aware network to adaptively bound the normalized range of each network block parameter, our approach improves the training speed by a 5 $\times$ compared to the standard Transformer architecture. In addition, this design has a large learning space compared to nGPT's method to ensure that the model can efficiently reconstruct haze features without violating the stability of training, which can be accelerated by deploying it on any mainstream GPU hardware.
ii) \textbf{Atmospheric Scattering-Aware} KV Caching: Motivated by the atmospheric scattering model~\cite{ju2017single, harris2003real}, we develop a key-value (KV) caching mechanism that dynamically optimizes feature retention based on haze density.  Unlike static attention mechanisms, this approach prioritizes features correlated with atmospheric scattering coefficients, thereby optimizing memory reuse during inference.  This strategy minimizes redundant computations while preserving critical scene details and prioritizing relevant contextual information in hazy regions. 
In addition, we propose anDehazeFormer, a method dedicated to image deinterpretability. 
Our primary contributions are summarized as follows:
\begin{itemize}
    \item We proposed a physically guided adaptive normalization mechanism. By dynamically calibrating the feature distribution with atmospheric parameters, the training speed is increased by five times while limiting the model complexity.
    \item For the first time, we integrated the physical prior of the haze into the KV caching process. Different from traditional caching mechanisms, which mainly depend on data statistical characteristics, this approach dynamically identifies the haze concentration distribution in images through a physical model and selectively retains key features related to haze removal.
    \item To improve the interpretability of haze removal models, we develop a visual attribution method that identifies and quantifies how specific regions or features contribute to model performance. Extensive experimental results demonstrate that our methods are best performers on publicly available benchmarks.
\end{itemize}

\section{Related work}%
\textbf{Single Image Dehazing.}
Single image dehazing has evolved from physics-based approaches to data-driven deep learning. Early methods like the Dark Channel Prior (DCP)~\cite{he2010single} estimated scene transmission using statistical priors but faced challenges in sky regions and high computation. Subsequent work, such as the Color Attenuation Prior (CAP)~\cite{dwivedi2023single}, linked depth to color saturation and brightness, yet remained constrained by manual priors~\cite{shao2020domain}. Deep learning revolutionized the field. DehazeNet~\cite{cai2016dehazenet} pioneered CNN-based transmission estimation, while AOD-Net~\cite{li2017aod} embedded atmospheric scattering models into end-to-end networks. Recent progress in image dehazing has seen attention-based approaches like FFA-Net~\cite{cheng2023ffa} and GridDehazeNet~\cite{liu2019griddehazenet} alongside unsupervised CycleGAN-based frameworks~\cite{yoo2020cyclegan}, with Wang et al.'s two-stage architecture achieving superior PSNR/SSIM through multi-scale attention and dark channel prior-guided haze component disentanglement~\cite{wang2025tsid}. 
Learning-based UHD image restoration has received some attention in recent years. While deep learning excels in conventional image restoration~\cite{cai2023retinexformer,cui2024revitalizing,wang2024dap}, direct application to UHD (4K/8K) faces scalability challenges: dense pixels can lead to excessive GPU memory consumption and even cause overflow errors on high end GPUs. For transformer based UHD images restoration, two main methods are currently adopted. One approach is to segment the image into small pieces, but there is a boundary error~\cite{wang2023ultra}, as well as downsampling upsampling ~\cite{wang2024correlation}. This reduces the computational load, but it may lose high-frequency details. Neither of them can integrate multi resolution information. It is worth noting that Li et al. ~\cite{li2023embedding} utilized frequency-domain similarity to restore low-light UHD images in the frequency domain,  but faces computational bottlenecks for real time applications.
Zheng et al.~\cite{zheng2021ultra} enabled real time 4K dehazing (125 fps) through low-resolution affine grid prediction and guided upsampling. In addition, to address the low resource 4K dehaze demand of mobile and IoT devices and lack of dataset, \cite{zheng20234k} proposed GAN simulation 4K haze map and construction of 4K-haze dataset. 
Despite progress, unresolved challenges persist, particularly the inefficiency in processing UHD images. It is necessary to study lightweight architectures and collaborative design of algorithms and hardware to accelerate. 

\textbf{The Key-Value (KV) cache mechanism.} The KV caching mechanism was originally designed to accelerate autoregressive generation in Transformers by reusing historical attention states, faces critical scalability challenges in UHD image processing due to excessive memory consumption and inefficient dynamic feature management.  Early approaches like $H_2O$~\cite{zhang2023h2o} dynamically evict low-impact tokens via entropy-based metrics, reducing KV cache memory by 95\%, Hu et al. allocate adaptive compression budgets across attention heads to maintain performance in long-context tasks~\cite{hu2024multi}. For UHD image restoration. However, these methods primarily target text generation and lack integration with physical priors critical for vision tasks. In image dehazing, GridDehazeNet embeds atmospheric scattering models into attention weights but retains static caching strategies~\cite{liu2019griddehazenet}. While Chen et al.~\cite{chen2025ultra} pioneered compressible attention caching inspired by LLM KV mechanisms, their approach overlooks the intrinsic physical constraints of haze distribution. Building on this foundation, we propose anDehazeFormer, KV cache dynamically prioritizes haze concentrated regions through transmission maps and global atmospheric light, enabling context-aware feature preservation.  
\section{Methodology}


This section presents the method details of anDehazeFormer, an innovative vision Transformer architecture integrating \textbf{a}daptive network layer weight \textbf{n}ormalization with a priori estimation of atmospheric physics and KV cache to address the problem of clarity in dehaze scenes with UHD resolution images. The method consists of three key components: an Atmospheric Parameter Estimator to model haze features, a KV Cache-Aware Window Attention module for efficient global context aggregation, and Atmospheric-Aware DehazeFormer Blocks to handle haze-sensitive features.  


\begin{figure}[t!]
    \centering
    \includegraphics[width=0.975\textwidth]{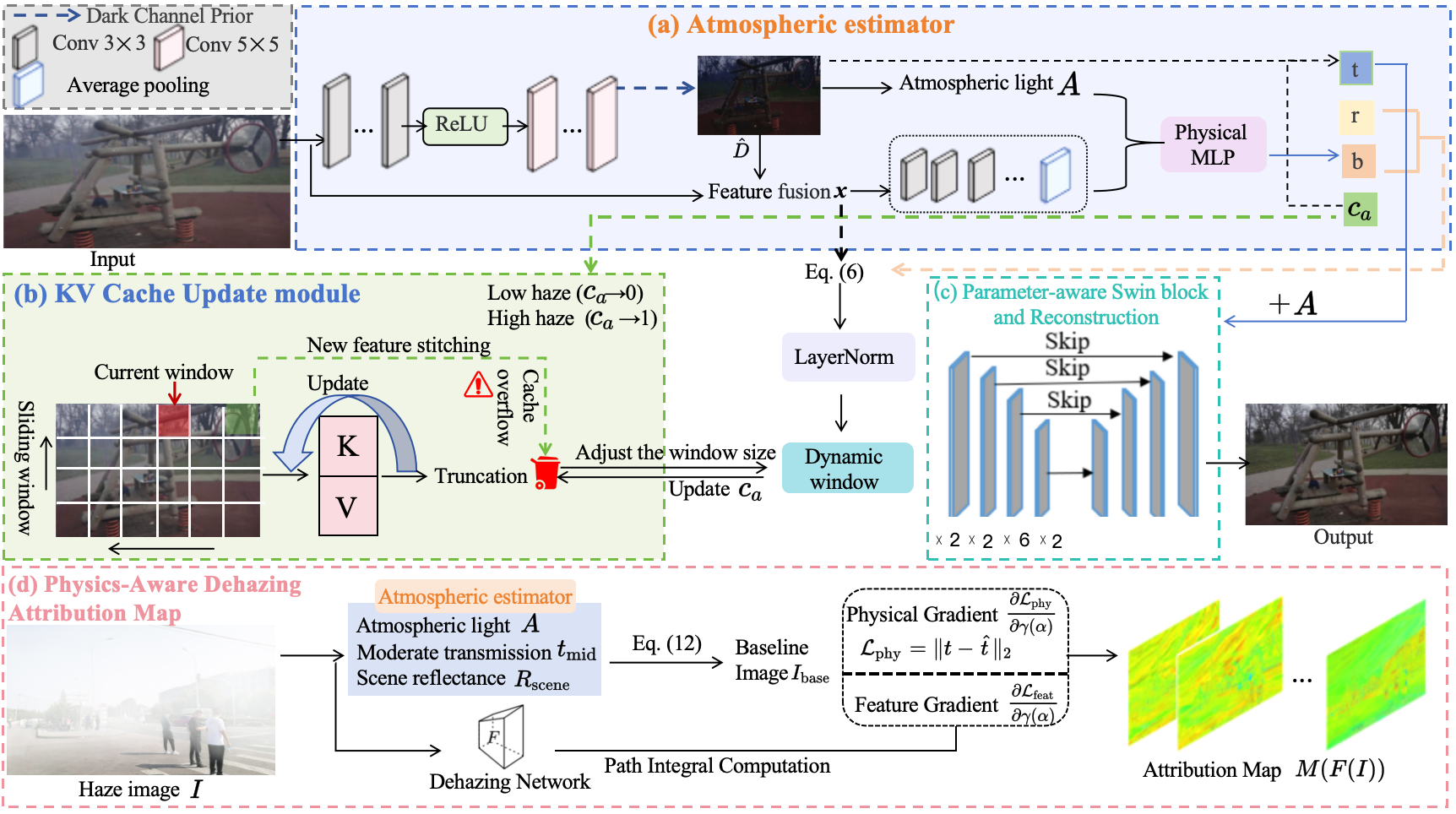}
    \caption{The framework integrates four key components.  (a) Atmospheric Estimator generating transmission maps and global light parameters, (b) KV Cache Update Module dynamically retaining features via atmospheric-guided retention policies, (c) Parameter-Aware Swin Blocks fusing physical constraints during multi-scale reconstruction, (d) Physics-Aware Attribution Map (PAAM) quantifying decision rationales through gradient integration. Modules are synergistically linked, Atmospheric parameters guide cache retention in (b) and feature restoration in (c), while PAAM (d) traces gradient flows from (a)-(c) to enhance interpretability.}
    \label{fig:model}
    \vspace{-1mm}
\end{figure}
\subsection{Atmospheric Parameter Estimator}
\label{subsec:atmospheric_estimator}

Our atmospheric parameter estimator establishes a crucial bridge between physical scattering models and deep feature learning, generating four essential parameters through a unified architecture: the transmission map $t$, atmospheric light $A$, and normalization factors $(r,b)$. As shown in Fig.~\ref{fig:model}(a), the module employs a dual-stream design that synergistically combines conventional atmospheric priors with learned feature representations.
Specifically, the dark channel computation forms the physical foundation of our estimator. Following the atmospheric scattering prior \cite{bruneton2008precomputed, kallweit2017deep}, we implement an efficient convolutional network to estimate the dark channel map:
\begin{equation}
    \hat{D} = \sigma\left(\text{Conv}_{3\times3}(\text{ReLU}(\text{Conv}_{5\times5}(x))\right), 
\end{equation}
where $\sigma(\cdot)$ denotes sigmoid activation. 
Atmospheric light estimation leverages this dark channel prior through a selective sampling process. We identify the brightest 0.1\% pixels in dark map $\hat{D}$ and compute atmospheric light $A$ as the mean RGB values at these locations. 
\begin{equation}
    A = \frac{1}{|\mathcal{T}|}\sum_{p\in\mathcal{T}}x(p),\ \mathcal{T}=\{p\ |\ \hat{D}(p)\geq\text{quantile}_{0.9\%}(\hat{D})\}.
\end{equation}
Then, the parameter fusion branch combines physical priors with deep feature learning. By connecting the original image $I$with the estimated dark channel $\hat{D}$, we can jointly process the original pixel data and physical constraints:
\begin{equation}
    f = \text{GlobalPool}\left(\text{Conv}_{3\times3}(\text{LeakyReLU}(\text{Conv}_{5\times5}(I \oplus \hat{D}))\right),
\end{equation}
where $\oplus$ denotes channel-wise concatenation. The resulting feature vector $f\in\mathbb{R}^d$ is concatenated with atmospheric light $A$ and processed through a constrained physical MLP. This design allows the network to learn adaptive corrections to the physical model based on global image context:
\begin{equation}
    [r, b, c_a] = \text{physical MLP}(f \oplus A).
\end{equation}
The final transmission map calculation explicitly encodes the atmospheric scattering model:
\begin{equation}
    t = 1 - c_a \odot \hat{D},
\end{equation}
where $\odot$ denotes element-wise multiplication. This formulation introduces a learnable scaling coefficient $c_a$ that adapts to varying haze densities and illumination conditions.


\subsection{Atmospheric Sensing KV Caching Mechanism Based on Adaptive Normalization}
\label{subsec:cache_transformer}

The cache-aware transformer block represents a significant architectural advancement that facilitates efficient global-local feature interaction while ensuring physical consistency. As illustrated in Fig.~\ref{fig:model}(b), varying haze distributions $c_a$ dynamically modulate the KV cache update strategy.

Inspired by the nGPT architecture, the scaling factor $r$ and the bias term $b$ are generated through the physical parameters (transmittance $t$, atmospheric light $A$) estimated by the atmospheric scattering model, and the physical characteristics of haze concentration are embedded into the normalization process:
\begin{equation}
    \text{LN}'(x) = \text{LayerNorm}(x \odot \text{Normal}(r) + \text{Normal}(b)).
    \label{eq: LN}
\end{equation}
Compared with the fixed parameters of traditional LayerNorm, this adaptive method allows the network to dynamically adjust the feature size based on haze density. Among them, a higher $r$ value amplifies the texture details of thick haze areas, while $b$ compensates for the color shift caused by atmospheric light, and is more adaptable to the differences in local haze distribution concentrations in UHD images.

Furthermore, due to the different resolutions of the input feature maps, we have designed a dynamic window to adaptively adjust the window size to avoid computational redundancy or information loss caused by a fixed window size. For example, when the number of windows increases at high resolution, the KV cache requires more efficient storage management to avoid memory overflow. For the input feature map with a size of $H \times W$, the window size $w_{\text{base}}$ is calculated as:
\begin{equation}
w_{\text{base}} = \underbrace{\left\lfloor \frac{\min(H, W)}{\alpha} \right\rfloor}_{\text{Basic size}} + \underbrace{\beta \cdot \mathbb{I}_{\{\min(H, W) > \tau\}}}_{\text{Ultra-high resolution compensation}},
\end{equation}
where, $\alpha=8$ controls the basic scale factor, $\beta=4$ provides the increment related to resolution, and $\tau=1024$ serves as the ultra-high resolution threshold. This formula balances the preservation of local details and computational efficiency.

\begin{algorithm}[t]
\caption{KV Cache Update with Atmospheric Guidance}
\begin{algorithmic}[1]
\State \textbf{Input:} Current keys $K_t$, values $V_t$, transmission coefficient $c_a$
\State \textbf{Persistent:} Key cache $\mathcal{K}$, value cache $\mathcal{V}$
\State \textbf{Hyperparameter:} Cache ratio $\eta$ (default=0.5)
\State \textcolor{gray}{\textit{// 1. Calculate retention ratio}}
\State $\gamma \gets 1 - \text{mean}(c_a) \cdot \eta$  
\State $n_{\text{keep}} \gets \lfloor \gamma \cdot |\mathcal{K}| \rfloor$

\State \textcolor{gray}{\textit{// 2. Feature alignment via linear interpolation}}
\If{$\mathcal{K} \neq \emptyset$ \textbf{and} $\text{dim}(K_t) \neq \text{dim}(\mathcal{K})$}
    \State $\tilde{K}_t \gets \text{Interpolate}(K_t, \text{size}=|\mathcal{K}|_L, \text{mode}=\text{linear})$
    \State $\tilde{V}_t \gets \text{Interpolate}(V_t, \text{size}=|\mathcal{K}|_L, \text{mode}=\text{linear})$
\Else
    \State $\tilde{K}_t \gets K_t,\ \tilde{V}_t \gets V_t$
\EndIf

\State \textcolor{gray}{\textit{// 3. Update cache with historical preservation}}
\State $\mathcal{K} \gets \text{Concat}(\mathcal{K}[:n_{\text{keep}}], \tilde{K}_t)$
\State $\mathcal{V} \gets \text{Concat}(\mathcal{V}[:n_{\text{keep}}], \tilde{V}_t)$

\State \textbf{Return} $\text{Attention}(Q, \mathcal{K}, \mathcal{V})$
\end{algorithmic}
\label{alg:cache_update}
\end{algorithm}

Since the dynamic window $w_{\text{base}}$ changes dynamically with the variation of resolution, the length of the KV sequence fluctuates. To maintain cache consistency, we propose an atmospheric scatter-aware KV cache optimization strategy (Algorithm \ref{alg:cache_update}), which integrates the physical characteristics of atmospheric scattering into the KV cache mechanism.
When the input resolution changes, causing the window size $w_{\text{base}}$ to be adjusted, the new key-value pairs $\mathbf{K}_t, \mathbf{V}_t$ may not match the sequence dimensions of the historical cache. To this end, the dimensional differences of the KV sequence are eliminated through interpolation alignment to ensure the compatibility of cross-resolution features. The aligned features will be physically and perceptually fused with the historical cache. Based on the transmittance $c_a$, the cache retention rate $\gamma = 1 - c_a \cdot \eta$ will be dynamically adjusted. Redundant features will be preferentially discarded in areas with high haze concentration, and $\eta$ is the preset cache attenuation rate.



\subsection{Atmospheric-Guided Feature Reconstruction}
\label{subsec:feature_pipeline}

Our method consists of two interdependent stages, parameterized feature encoding through Swin Transformer blocks and multi-scale reconstruction guided by atmospheric scattering principles.

\subsubsection{Atmospheric-Aware DehazeFormer Block}  
The Parameter-Aware Swin Transformer Block (PA-STB) fundamentally incorporates atmospheric parameters into the feature transformation process. In Eq.~\ref{eq: LN}, for input features \( x \in \mathbb{R}^{H \times W \times C} \), the normalized features are processed through cache-aware window attention with dynamic window partitioning. When the input resolution \( (H, W) \) varies, the window size automatically adjusts to \( w_{\text{adapt}} = \min(w_{\text{base}}, H, W) \). Within each \( w_{\text{adapt}} \times w_{\text{adapt}} \) window, the attention mechanism effectively integrates local details with cached global context:
\[
x_{attn} = \text{Softmax}\left(\frac{Q[K_{\text{cache}};K]^\top}{\sqrt{d}} + \hat{B}\right)[V_{\text{cache}};V],
\]  
where $K_{\text{cache}},V_{\text{cache}}$ retain historical patterns guided by transmission parameter $c_a$. The attention output is then refined through a parameterized residual connection:  
\[
x_{\text{out}} = x + \text{DropPath}(\gamma_1 \odot x_{attn} + \beta_1).
\]  
Finally, a bottleneck MLP with hidden dimension $4C$ processes the features:  
\[
x_{\text{final}} = x_{\text{out}} + \text{DropPath}(\gamma_2 \odot \text{MLP}(\text{LayerNorm}(x_{\text{out}})) + \beta_2),
\]  
where $\gamma_{1:2}, \beta_{1:2} \in \mathbb{R}^C$ stabilize gradient flow. 


\subsubsection{Multi-Scale Reconstruction}
\label{subsec:multi_scale}

The multi-scale reconstruction module employs a hybrid architecture that combines progressive feature refinement with resolution recovery, as illustrated in Fig.~\ref{fig:model}(c). The process begins with the compressed feature map $\mathcal{F}_d \in \mathbb{R}^{C\times H'\times W'}$ from the transformer backbone, where $(H', W') = (\frac{H}{16}, \frac{W}{16})$ for initial 256px inputs. Through cascaded convolution and interpolation operations, we restore both spatial resolution and texture details while maintaining atmospheric consistency.


The reconstruction pipeline consists of three core phases, First, depth-wise separable convolutions enhance features:
\begin{equation}
    \mathcal{F}_{\text{refined}} = \text{DWConv}_{3\times3}(\text{ReLU}(\text{BN}(\text{Conv}_{1\times1}(\mathcal{F}_d)))).
\end{equation}
DWConv reduces parameters by 38\%  compared to standard convolutions. Bilinear interpolation upsamples features with physical constraints, explicitly embeds the atmospheric scattering model into the upsampling process:
\begin{equation}
    \mathcal{F}_{\text{up}} = \text{Interp}(\mathcal{F}_{\text{refined}}, 4\times) \odot t_{\uparrow} + A_{\uparrow} \odot (1 - t_{\uparrow}),
\end{equation}
where $t_{\uparrow}$ and $A_{\uparrow}$ are resized versions of the estimated transmission map and atmospheric light, respectively. This formulation explicitly embeds the atmospheric scattering model during reconstruction. This explicit integration of the atmospheric scattering model prevents color distortion during upsampling. The final reconstruction combines enhanced features with original image residuals:
\begin{equation}
    \hat{J} = \text{Conv}_{3\times3}(\mathcal{F}_{\text{up}}) + I_{\text{orig}}.
\end{equation}
The final output combines three critical elements: High-frequency details from the refined features, Atmospheric physics constraints through $t$ and $A$, and Original image residuals for color preservation. 


\subsection{Physics-Aware Dehazing Attribution Map}
\begin{figure}[t]
    \centering
    \caption{Illustration of the baseline image and the path function.}
    \includegraphics[width=0.95\textwidth]{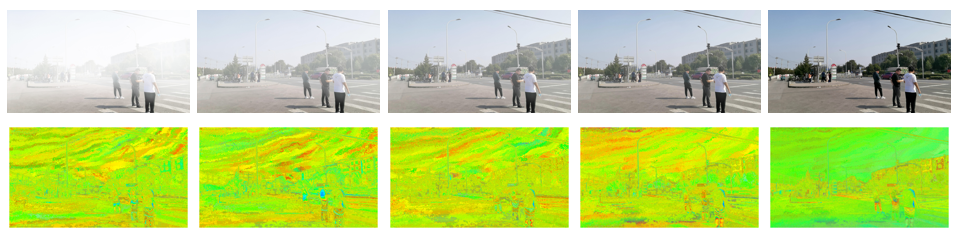}
    \label{fig:ima}
\end{figure}

Inspired by the Dehazing Attribution Map (DAM)~\cite{chen2025tokenize0} and Atmospheric Scattering Model (ASM)~\cite{ju2021ide}, to enhance model transparency, we propose a Physics-Aware Attribution Map (PAAM) that enhances the interpretability of dehazing networks by integrating gradients from both atmospheric scattering models and deep features. Let $\textbf{F}$ represent a dehazing network, for a hazy image $I$, to conduct attribution analysis for the dehazing network, we require a baseline input image $ I_{\text{base}}$ that satisfies that $\textbf{F}( I_{\text{base}})$ is absent certain features present in $\textbf{F}(I)$. As shown in Fig.~\ref{fig:ima}, the Physics-Aware Attribution Map $\textbf{M}(\textbf{F}(I))$ is obtained by computing the path integral of physical and feature gradients. This smooth path function is denoted as \( \gamma(\alpha): [0,1] \to \mathbb{R}^{h \times w} \), with \( \gamma(0) : I' \) and \( \gamma(1) : I \).
\begin{equation}
   \textbf{M}(\textbf{F}(I)) = \int_0^1  \frac{\partial \mathcal{L}_{\text{phy}}(\textbf{F}( I_{\text{base}}))}{\partial \gamma(\alpha)} \cdot \lambda \frac{\partial \mathcal{L}_{\text{feat}}(\textbf{F}( I_{\text{base}}))}{\partial \gamma(\alpha)}  d\alpha, 
\end{equation}
where \(\mathcal{L}_{\text{phy}} = \| t - \hat{t} \|_2\) is the transmission estimation loss, \(\mathcal{L}_{\text{feat}}\) denotes feature reconstruction loss. 

Previous studies have emphasized that the validity of model attribution essentially depends on the selection of an appropriate baseline~\cite{sturmfels2020visualizing}. In the task of image classification, pure black images are usually taken as the baseline. However, for the task of removing haze, it is more appropriate to choose clear images as the baseline because the restoration of haze areas is a key challenge. The attribution analysis of the haze removal network needs to satisfy dual constraints: 1) The baseline image needs to retain the basic structure of the scene; 2) The path evolution needs to reflect the continuous changes in haze concentration. In this work, we elaborately designed a physically guided baseline generation strategy for the haze removal network. We utilize the $I_{\text{base}}$ as the baseline input.
\begin{equation}
    I_{\text{base}} = A \odot (1 - t_{\text{mid}}) + R_{\text{scene}},\quad t_{\text{mid}} = 0.7,
\end{equation}
where $R_{\text{scene}}$ is the scene reflectance estimated, which simulates the color of objects under conditions without haze. $t_{\text{mid}}$ represents moderate transmission, indicating that the haze concentration in the baseline image is moderate. This not only retains the scene structure but also injects controllable haze, avoiding the information bias of a completely black or fully clear baseline. 
\section{Experiments and Analysis}
\subsection{Dataset and Implementation Details}
\textbf{Dataset.} To train and evaluate our proposed method, we used multiple datasets. The 8KDehaze dataset~\cite{chen2025tokenize0} includes 100 UHD images at 8192$\times$8192 pixels, derived from satellite imagery and professional-grade photographs, with haze synthesized through a refined atmospheric scattering model that accounts for humidity and particulate matter density. To address synthetic diversity, the 4KID dataset~\cite{zheng2021ultra} offers computationally generated scenes featuring varied textures, lighting conditions, and object types. 
Real-world performance is evaluated using two datasets: O-HAZE~\cite{Oancuti2018haze} includes 45 outdoor hazy/clean pairs (1286$\times$947–5436$\times$3612) captured with physical haze machines in natural environments to test texture/edge preservation under complex haze;  I-HAZE~\cite{Iancuti2018haze} has 35 indoor pairs (2K–4K) generated via standardized haze rooms to evaluate color fidelity/scene consistency under uniform haze. This multi-resolution, multi-scenario design ensures rigorous testing across synthetic complexity, high-resolution challenges, and practical real-world applications.

\textbf{Implementation Details.}
The proposed framework was implemented using PyTorch 2.1.0 and trained on an NVIDIA RTX 4090D GPU (24GB VRAM).  
For comparative analysis, we selected a series of the most advanced haze removal algorithms published recently. Including 4KDehazing~\cite{zheng2021ultra}, Dehamer~\cite{Guo_2022_CVPR}, DehazeFormer~\cite{song2023vision}, MB-TaylorFormer~\cite {Qiu_2023_ICCV}, DEA-Net~\cite{chen2024dea} and DehazeXL~\cite{chen2025tokenize0}. Since these methods cannot be directly trained on 4K and 8K images, the image resolution is adjusted to 910$\times$512. Then the input image pairs are randomly cropped into patches of size 512$\times$512 pixels, with a batch size of 4 to accommodate GPU memory constraints.  We adopted the Adam optimizer with an initial learning rate of 0.001 and weight decay of 1e-4, employing a cosine annealing schedule for learning rate decay over 500 training epochs.  The loss function combines L1 reconstruction loss (0.8 weight), MSE loss (0.1 weight), and SSIM structural similarity loss (0.1 weight) to balance pixel-level accuracy and perceptual quality. Training convergence required approximately 12 hours on our hardware setup. During inference, the full framework processes 512$\times$512 images at 53 FPS, demonstrating real-time capability for 4K resolution (3840$\times$2160) through sliding window implementation. We adopted an 80-10-10 data split strategy for all experiments, allocating 80\% of the data for training, 10\% for validation, and 10\% for testing.

\subsection{Evaluation and Results}
\begin{figure}[h]
    \centering
    \includegraphics[width=\linewidth]{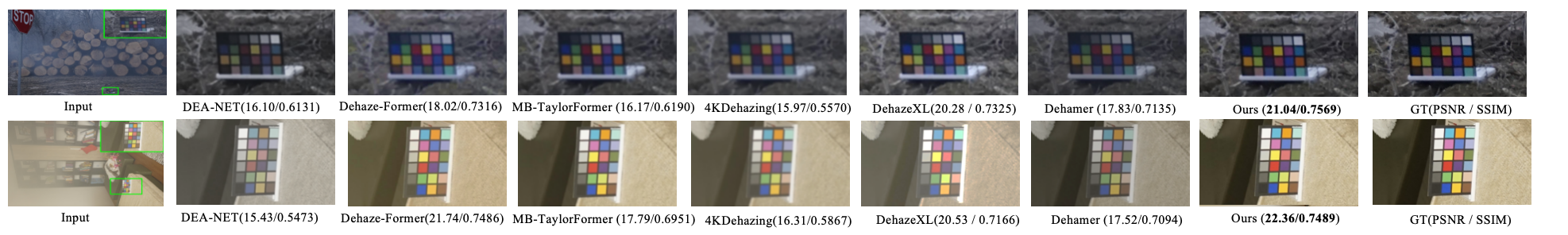}
    \caption{Dehazed results on the O-HAZE~\cite{Oancuti2018haze} and I-HAZE~\cite{Iancuti2018haze} dataset. The proposed anDehazeFormer demonstrates higher color fidelity and restores more details compared with other SOTA methods.}
    \label{fig:IO-HAZE}
\end{figure}

\begin{figure}[h]
    \centering
    \includegraphics[width=\linewidth]{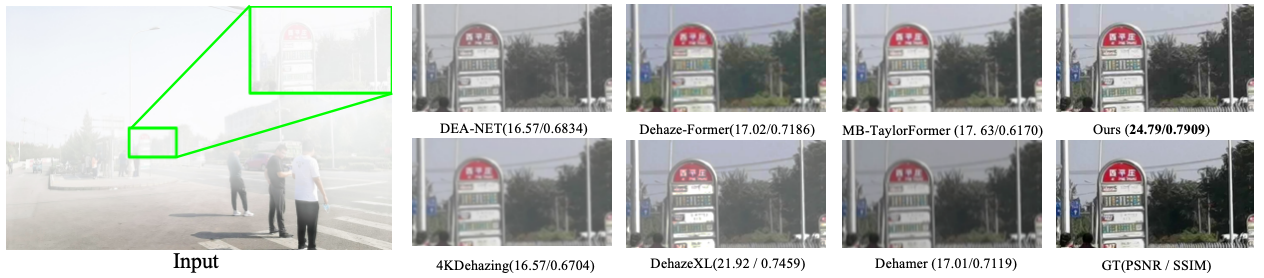}
    \caption{Dehazed results on the  4KID~\cite{zheng2021ultra} dataset. The proposed anDehazeFormer effectively avoids ghosting and blurring, achieving full-resolution inference with remarkable clarity and precision.}
    \label{fig:4KID}
\end{figure}

\begin{figure}[h]
    \centering
    \includegraphics[width=\linewidth]{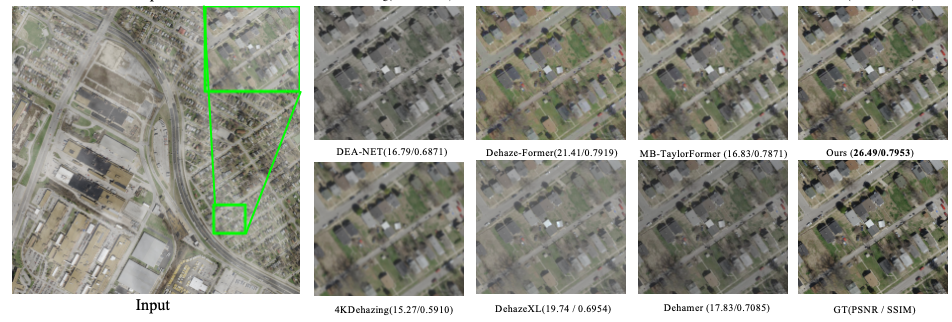}
    \caption{Dehazed results on the  8KDehaze~\cite{chen2025tokenize0} dataset. The proposed anDehazeFormer effectively eliminates segmentation artifacts and achieves superior visual quality.}
    \label{fig:8KDehaze}
\end{figure}

\begin{figure}[h]
    \centering
    \includegraphics[width=\linewidth]{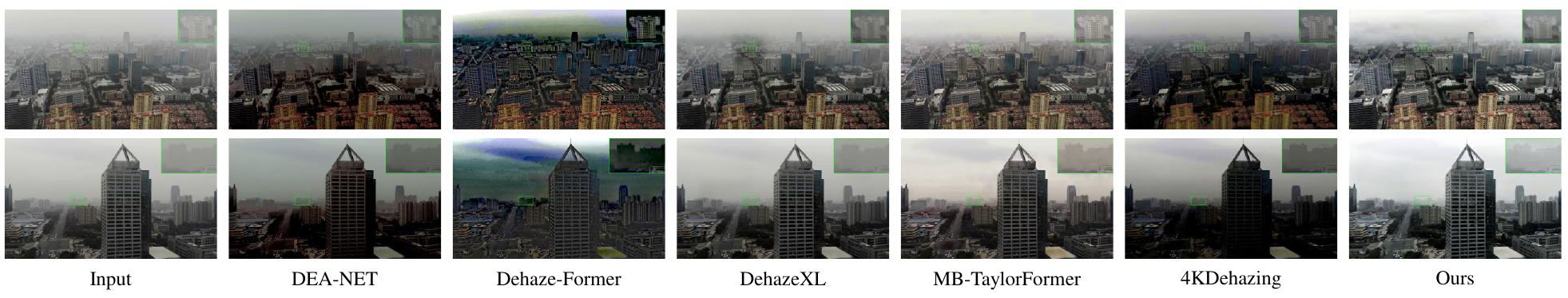}
    \caption{Dehazed results in the real world.}
    \label{fig:real}
\end{figure}

\textbf{Qualitative Evaluation.}
Figure~\ref{fig:IO-HAZE} to~\ref{fig:real} present the visual results of the proposed method and comparative algorithms.
As illustrated in Figure.~\ref{fig:IO-HAZE}, 4KDehazing~\cite{zheng2021ultra} causes local supersaturation, MB-TaylorFormer~\cite{Qiu_2023_ICCV} introduces blurring, especially in vegetation and architecture details. The anDehazeFormer dynamically detects haze concentration distribution via a physical model, accurately restores high frequency details and color fidelity.
In Figure~\ref{fig:4KID}, the existing haze removal methods generally have performance degradation in the sky area. Due to the high similarity in spectral distribution between the low-contrast feature of the natural sky and the haze scattering effect, the slicing reasoning method is difficult to effectively distinguish the real haze from the characteristics of the sky area. In contrast, the anDehazeFormer embeds the physical prior depth into the neural network architecture, effectively distinguishing between the sky and haze regions, boosting PSNR to 24.79 dB and SSIM to 0.7909, 2.87 dB and 0.045 above the suboptimal method DehazeXL~\cite{chen2025tokenize0}. 
Figure~\ref{fig:8KDehaze} further emphasizes the advantages of anDehazeFormer in color restoration and overall coherence.

As shown in Figure~\ref{fig:real}, in order to evaluate the generalization ability, we conducted experiments on two randomly captured 4K resolution real hazy images (3840×2160). Dehaze-Former~\cite{song2023vision} shows obvious color shift and supersaturation in the distant area, while our method restores fine texture more effectively while maintaining color fidelity, demonstrating superior haze removal performance.


\textbf{Quantitative Evaluation.}
Table~\ref{tab:1} presents the quantitative evaluation outcomes of anDehazeFormer and the compared methods on the 8KDehaze~\cite{chen2025tokenize0}, 4KID~\cite{zheng2021ultra}, O-HAZE~\cite{Oancuti2018haze}, and I-HAZE~\cite{Iancuti2018haze} datasets, employing metrics like PSNR, SSIM~\cite{wang2004image}, and average inference time. The proposed method achieves the highest scores for both PSNR and SSIM, indicating its superior dehazing effectiveness. Although 4KDehazing~\cite{zheng2021ultra} is close to our method in terms of computational speed, it has significant performance degradation when processing large-scale images (such as 8K resolution), accompanied by ghosting and color shift issues. In contrast, anDehazeFormer achieves a good balance between haze removal performance and processing time, on the 8KDehaze~\cite{chen2025tokenize0} dataset, the PSNR/SSIM metric is 13.48dB/0.124 higher than that of 4KDehazing~\cite{zheng2021ultra}.
These results validate anDehazeFormer as a practical solution for real-world dehazing tasks, effectively bridging the gap between computational efficiency and high-quality restoration.

Furthermore, this study compared the FLOPs of the mainstream models. The FLOPs of this model is only 15.8G, significantly lower than methods such as 4KDehazing~\cite{zheng2021ultra} (77.29.5G) and Dehamer~\cite{Guo_2022_CVPR} (120.6G). In addition, our model only takes 1.24s on 8K images, which is approximately 3.7 times faster than DehazeXL~\cite{chen2025tokenize0}, which has a similar FLOPs (4.62s)

\begin{table}[h]
    \centering
    \caption{Quantitative evaluations on the 8KDehaze dataset~\cite{chen2025tokenize0}, the 4KID dataset~\cite{zheng2021ultra}, the O-HAZE dataset~\cite{Oancuti2018haze} and the I-HAZE~\cite{Iancuti2018haze} in terms of PSNR, SSIM, and average infer time.}
    \label{tab:1}
    \resizebox{\textwidth}{!}{
    \begin{tabular}{llcccccccccccc}
        \toprule
        \multirow{2}{*}{Method} & \multirow{2}{*}{Venue\&Year} & \multicolumn{3}{c}{8KDehaze~\cite{chen2025tokenize0}} & \multicolumn{3}{c}{4KID~\cite{zheng2021ultra}} & \multicolumn{3}{c}{O-HAZY~\cite{Oancuti2018haze}} & \multicolumn{3}{c}{I-HAZE~\cite{Iancuti2018haze}} \\
         & & PSNR & SSIM & Time(s) & PSNR & SSIM & Time(s) & PSNR & SSIM & Time(s) & PSNR & SSIM & Time(s) \\
        \midrule
        4KDehazing~\cite{zheng2021ultra} & CVPR2021 & 20.41 & 0.8664 & 1.35 & 18.68 & 0.7424 & 0.19 & 19.30 & 0.6426 & 0.27 & 18.43 & 0.6537 & 0.23
        \\
         Dehamer~\cite{Guo_2022_CVPR} &  CVPR2022 &  25.92 &  0.9373 & 6.61 & 21.24 &  0.8795 & 1.03 & 19.59 & 0.7134 &1.30 & 18.82 & 0.6731 & 1.32
        \\
        DehazeFormer~\cite{song2023vision} & TIP2023 &  26.68 & 0.9729 & 7.47 &  20.83 & 0.8763 & 1.17 &  19.86 &  0.7116 & 1.47 & 19.04 & 0.6598 & 1.51
        \\
         MB-TaylorFormer~\cite{Qiu_2023_ICCV} & ICCV2023 & 26.41 & 0.9668 & 120.54 & 18.63 &  0.8497 & 18.83 & 19.57 & 0.7104 & 23.71 & 18.87 & 0.6691 & 24.57
        \\
         DEA-Net~\cite{chen2024dea} & TIP2024 &  25.89 & 0.9329 & 7.40 & 20.83 & 0.8834 & 1.16 &  20.01 & 0.6988 & 1.46 & 19.88 & 0.6780 & 1.49
        \\
        DehazeXL~\cite{chen2025tokenize0} & CVPR2025 &  32.35 & 0.9863 & 4.62 & 26.62 &  0.9073 & 0.59 & 21.49 &  0.7348 & 0.86 & 20.81 & 0.7051 & 0.88
        \\
        \midrule
        \multicolumn{2}{c}{Ours}  & 33.89 & 0.9899 & 1.24 & 27.85 & 0.9177 & 0.22 & 22.85 & 0.7333 & 0.21 & 20.88 & 0.7224 & 0.19
        \\
        \bottomrule
    \end{tabular}}
\end{table}

\subsection{Ablation Study}
We conduct comprehensive ablation experiments to validate the effectiveness of key components in anDehazeFormer. All experiments are performed on the 4KID dataset under identical training settings (batch size = 4,500 epochs), with the results presented in Table~\ref{tab:ablation}. The adaptive normalization layer stabilizes training by dynamically adjusting features based on haze characteristics, while the atmospheric-guided KV cache optimizes memory usage through physics-aware feature retention. The multi-scale reconstruction module, integrating transmission maps and atmospheric constraints, ensures detail preservation in dense haze regions. 
\begin{table}[h]
    \centering
    \caption{Ablation study on 4KID dataset (512$\times$512 crops)}
    \label{tab:ablation}
    \begin{tabular}{lcccc}
        \toprule
        Configuration & PSNR↑ & SSIM↑ & Memory (GB)↓ & FPS↑ \\
        \midrule
        Baseline & 24.71 & 0.8632 & 18.3 & 34 \\
        + Adaptive Norm & 26.28 & 0.8915 & 18.5 & 41 \\
        + KV Cache & 27.09 & 0.9037 & 14.2 & 49 \\
        + Multi-Scale Reconstruction & 25.36 & 0.8709 & 17.4 & 38 \\
        Full Model & 28.45 & 0.9217 & 15.6 & 53 \\
        \bottomrule
    \end{tabular}
\end{table}
\section{Conclusion and Limitation}
\textbf{Conclusion. }We present anDehazeFormer, a transformer-based framework for ultra-high-resolution image dehazing that integrates adaptive normalization and atmospheric-guided KV caching.  The adaptive normalization stabilizes training by dynamically adjusting features to haze conditions, while the physics-aware KV cache optimizes memory usage through context-aware feature retention.  

\textbf{Limitation. }Although this method effectively enhances the real-time processing capability of UHD haze removal by introducing the KV caching mechanism, with the increase of network depth, its storage overhead grows linearly with the model depth and feature dimension, which may pose challenges to the hardware resources of large-scale scene applications (such as ultra-high-resolution video stream processing). This limitation stems from the layer-by-layer accumulation characteristic of key-value pairs in the self-attention mechanism, especially when dealing with long sequence inputs, the problem of video memory occupation is more significant. In the future, we will study the structured compression of the KV cache matrix based on the low-rank decomposition theory and utilize the redundancy of the feature space to reduce the storage requirements.
\bibliography{egbib}


\newpage
\section*{NeurIPS Paper Checklist}


\begin{enumerate}

\item {\bf Claims}
    \item[] Question: Do the main claims made in the abstract and introduction accurately reflect the paper's contributions and scope?
    \item[] Answer: \answerYes{} 
    \item[] Justification: The main claims made in the abstract and introduction accurately reflect this paper’s contributions and scope.
    \item[] Guidelines:
    \begin{itemize}
        \item The answer NA means that the abstract and introduction do not include the claims made in the paper.
        \item The abstract and/or introduction should clearly state the claims made, including the contributions made in the paper and important assumptions and limitations. A No or NA answer to this question will not be perceived well by the reviewers. 
        \item The claims made should match theoretical and experimental results, and reflect how much the results can be expected to generalize to other settings. 
        \item It is fine to include aspirational goals as motivation as long as it is clear that these goals are not attained by the paper. 
    \end{itemize}

\item {\bf Limitations}
    \item[] Question: Does the paper discuss the limitations of the work performed by the authors?
    \item[] Answer: \answerYes{} 
    \item[] Justification: Although this method effectively enhances the real-time processing capability of UHD haze removal by introducing the KV caching mechanism, with the increase of network depth, its storage overhead grows linearly with the model depth and feature dimension, which may pose challenges to the hardware resources of large-scale scene applications.
    \item[] Guidelines:
    \begin{itemize}
        \item The answer NA means that the paper has no limitation while the answer No means that the paper has limitations, but those are not discussed in the paper. 
        \item The authors are encouraged to create a separate "Limitations" section in their paper.
        \item The paper should point out any strong assumptions and how robust the results are to violations of these assumptions (e.g., independence assumptions, noiseless settings, model well-specification, asymptotic approximations only holding locally). The authors should reflect on how these assumptions might be violated in practice and what the implications would be.
        \item The authors should reflect on the scope of the claims made, e.g., if the approach was only tested on a few datasets or with a few runs. In general, empirical results often depend on implicit assumptions, which should be articulated.
        \item The authors should reflect on the factors that influence the performance of the approach. For example, a facial recognition algorithm may perform poorly when image resolution is low or images are taken in low lighting. Or a speech-to-text system might not be used reliably to provide closed captions for online lectures because it fails to handle technical jargon.
        \item The authors should discuss the computational efficiency of the proposed algorithms and how they scale with dataset size.
        \item If applicable, the authors should discuss possible limitations of their approach to address problems of privacy and fairness.
        \item While the authors might fear that complete honesty about limitations might be used by reviewers as grounds for rejection, a worse outcome might be that reviewers discover limitations that aren't acknowledged in the paper. The authors should use their best judgment and recognize that individual actions in favor of transparency play an important role in developing norms that preserve the integrity of the community. Reviewers will be specifically instructed to not penalize honesty concerning limitations.
    \end{itemize}

\item {\bf Theory assumptions and proofs}
    \item[] Question: For each theoretical result, does the paper provide the full set of assumptions and a complete (and correct) proof?
    \item[] Answer:\answerNA{} 
    \item[] Justification: The paper does not include theoretical results.
    \item[] Guidelines:
    \begin{itemize}
        \item The answer NA means that the paper does not include theoretical results. 
        \item All the theorems, formulas, and proofs in the paper should be numbered and cross-referenced.
        \item All assumptions should be clearly stated or referenced in the statement of any theorems.
        \item The proofs can either appear in the main paper or the supplemental material, but if they appear in the supplemental material, the authors are encouraged to provide a short proof sketch to provide intuition. 
        \item Inversely, any informal proof provided in the core of the paper should be complemented by formal proofs provided in appendix or supplemental material.
        \item Theorems and Lemmas that the proof relies upon should be properly referenced. 
    \end{itemize}

    \item {\bf Experimental result reproducibility}
    \item[] Question: Does the paper fully disclose all the information needed to reproduce the main experimental results of the paper to the extent that it affects the main claims and/or conclusions of the paper (regardless of whether the code and data are provided or not)?
    \item[] Answer: \answerYes{} 
    \item[] Justification: We describe our proposed anDehazeFormer clearly and fully in Section 3.1, Section 3.2 Section 3.3 and Section 3.4.   We illustrate our experimental setting and details in Section 4.1, Section 4.2 and  Section 4.4. Our code will be made public after the article is published.
    \item[] Guidelines:
    \begin{itemize}
        \item The answer NA means that the paper does not include experiments.
        \item If the paper includes experiments, a No answer to this question will not be perceived well by the reviewers: Making the paper reproducible is important, regardless of whether the code and data are provided or not.
        \item If the contribution is a dataset and/or model, the authors should describe the steps taken to make their results reproducible or verifiable. 
        \item Depending on the contribution, reproducibility can be accomplished in various ways. For example, if the contribution is a novel architecture, describing the architecture fully might suffice, or if the contribution is a specific model and empirical evaluation, it may be necessary to either make it possible for others to replicate the model with the same dataset, or provide access to the model. In general. releasing code and data is often one good way to accomplish this, but reproducibility can also be provided via detailed instructions for how to replicate the results, access to a hosted model (e.g., in the case of a large language model), releasing of a model checkpoint, or other means that are appropriate to the research performed.
        \item While NeurIPS does not require releasing code, the conference does require all submissions to provide some reasonable avenue for reproducibility, which may depend on the nature of the contribution. For example
        \begin{enumerate}
            \item If the contribution is primarily a new algorithm, the paper should make it clear how to reproduce that algorithm.
            \item If the contribution is primarily a new model architecture, the paper should describe the architecture clearly and fully.
            \item If the contribution is a new model (e.g., a large language model), then there should either be a way to access this model for reproducing the results or a way to reproduce the model (e.g., with an open-source dataset or instructions for how to construct the dataset).
            \item We recognize that reproducibility may be tricky in some cases, in which case authors are welcome to describe the particular way they provide for reproducibility. In the case of closed-source models, it may be that access to the model is limited in some way (e.g., to registered users), but it should be possible for other researchers to have some path to reproducing or verifying the results.
        \end{enumerate}
    \end{itemize}

\item {\bf Open access to data and code}
    \item[] Question: Does the paper provide open access to the data and code, with sufficient instructions to faithfully reproduce the main experimental results, as described in supplemental material?
    \item[] Answer: \answerYes{} 
    \item[] Justification:  Our code will be obtained at \url{https://anonymous.4open.science/r/anDehazeFormer-632E/README.md}.
    \item[] Guidelines:
    \begin{itemize}
        \item The answer NA means that paper does not include experiments requiring code.
        \item Please see the NeurIPS code and data submission guidelines (\url{https://nips.cc/public/guides/CodeSubmissionPolicy}) for more details.
        \item While we encourage the release of code and data, we understand that this might not be possible, so “No” is an acceptable answer. Papers cannot be rejected simply for not including code, unless this is central to the contribution (e.g., for a new open-source benchmark).
        \item The instructions should contain the exact command and environment needed to run to reproduce the results. See the NeurIPS code and data submission guidelines (\url{https://nips.cc/public/guides/CodeSubmissionPolicy}) for more details.
        \item The authors should provide instructions on data access and preparation, including how to access the raw data, preprocessed data, intermediate data, and generated data, etc.
        \item The authors should provide scripts to reproduce all experimental results for the new proposed method and baselines. If only a subset of experiments are reproducible, they should state which ones are omitted from the script and why.
        \item At submission time, to preserve anonymity, the authors should release anonymized versions (if applicable).
        \item Providing as much information as possible in supplemental material (appended to the paper) is recommended, but including URLs to data and code is permitted.
    \end{itemize}

\item {\bf Experimental setting/details}
    \item[] Question: Does the paper specify all the training and test details (e.g., data splits, hyperparameters, how they were chosen, type of optimizer, etc.) necessary to understand the results?
    \item[] Answer: \answerYes{} 
    \item[] Justification: We illustrate our experimental setting and details in Section 4.1.
    \item[] Guidelines:
    \begin{itemize}
        \item The answer NA means that the paper does not include experiments.
        \item The experimental setting should be presented in the core of the paper to a level of detail that is necessary to appreciate the results and make sense of them.
        \item The full details can be provided either with the code, in appendix, or as supplemental material.
    \end{itemize}

\item {\bf Experiment statistical significance}
    \item[] Question: Does the paper report error bars suitably and correctly defined or other appropriate information about the statistical significance of the experiments?
    \item[] Answer: \answerYes{} 
    \item[] Justification: Table 1 shows the experimental results on different methods.
    \item[] Guidelines:
    \begin{itemize}
        \item The answer NA means that the paper does not include experiments.
        \item The authors should answer "Yes" if the results are accompanied by error bars, confidence intervals, or statistical significance tests, at least for the experiments that support the main claims of the paper.
        \item The factors of variability that the error bars are capturing should be clearly stated (for example, train/test split, initialization, random drawing of some parameter, or overall run with given experimental conditions).
        \item The method for calculating the error bars should be explained (closed form formula, call to a library function, bootstrap, etc.)
        \item The assumptions made should be given (e.g., Normally distributed errors).
        \item It should be clear whether the error bar is the standard deviation or the standard error of the mean.
        \item It is OK to report 1-sigma error bars, but one should state it. The authors should preferably report a 2-sigma error bar than state that they have a 96\% CI, if the hypothesis of Normality of errors is not verified.
        \item For asymmetric distributions, the authors should be careful not to show in tables or figures symmetric error bars that would yield results that are out of range (e.g. negative error rates).
        \item If error bars are reported in tables or plots, The authors should explain in the text how they were calculated and reference the corresponding figures or tables in the text.
    \end{itemize}

\item {\bf Experiments compute resources}
    \item[] Question: For each experiment, does the paper provide sufficient information on the computer resources (type of compute workers, memory, time of execution) needed to reproduce the experiments?
    \item[] Answer: \answerYes{} 
    \item[] Justification: The proposed framework was implemented using PyTorch 2.1.0 and trained on an NVIDIA RTX 4090D GPU (24GB VRAM).
    \item[] Guidelines:
    \begin{itemize}
        \item The answer NA means that the paper does not include experiments.
        \item The paper should indicate the type of compute workers CPU or GPU, internal cluster, or cloud provider, including relevant memory and storage.
        \item The paper should provide the amount of compute required for each of the individual experimental runs as well as estimate the total compute. 
        \item The paper should disclose whether the full research project required more compute than the experiments reported in the paper (e.g., preliminary or failed experiments that didn't make it into the paper). 
    \end{itemize}
    
\item {\bf Code of ethics}
    \item[] Question: Does the research conducted in the paper conform, in every respect, with the NeurIPS Code of Ethics \url{https://neurips.cc/public/EthicsGuidelines}?
    \item[] Answer: \answerYes{} 
    \item[] Justification: This research conforms in every respect with the NeurIPS Code of Ethics.
    \item[] Guidelines:
    \begin{itemize}
        \item The answer NA means that the authors have not reviewed the NeurIPS Code of Ethics.
        \item If the authors answer No, they should explain the special circumstances that require a deviation from the Code of Ethics.
        \item The authors should make sure to preserve anonymity (e.g., if there is a special consideration due to laws or regulations in their jurisdiction).
    \end{itemize}

\item {\bf Broader impacts}
    \item[] Question: Does the paper discuss both potential positive societal impacts and negative societal impacts of the work performed?
    \item[] Answer:\answerYes{} 
    \item[] Justification: Have no social influence
    \item[] Guidelines:
    \begin{itemize}
        \item The answer NA means that there is no societal impact of the work performed.
        \item If the authors answer NA or No, they should explain why their work has no societal impact or why the paper does not address societal impact.
        \item Examples of negative societal impacts include potential malicious or unintended uses (e.g., disinformation, generating fake profiles, surveillance), fairness considerations (e.g., deployment of technologies that could make decisions that unfairly impact specific groups), privacy considerations, and security considerations.
        \item The conference expects that many papers will be foundational research and not tied to particular applications, let alone deployments. However, if there is a direct path to any negative applications, the authors should point it out. For example, it is legitimate to point out that an improvement in the quality of generative models could be used to generate deepfakes for disinformation. On the other hand, it is not needed to point out that a generic algorithm for optimizing neural networks could enable people to train models that generate Deepfakes faster.
        \item The authors should consider possible harms that could arise when the technology is being used as intended and functioning correctly, harms that could arise when the technology is being used as intended but gives incorrect results, and harms following from (intentional or unintentional) misuse of the technology.
        \item If there are negative societal impacts, the authors could also discuss possible mitigation strategies (e.g., gated release of models, providing defenses in addition to attacks, mechanisms for monitoring misuse, mechanisms to monitor how a system learns from feedback over time, improving the efficiency and accessibility of ML).
    \end{itemize}
    
\item {\bf Safeguards}
    \item[] Question: Does the paper describe safeguards that have been put in place for responsible release of data or models that have a high risk for misuse (e.g., pretrained language models, image generators, or scraped datasets)?
    \item[] Answer: \answerNA{} 
    \item[] Justification: This paper poses no such risks.
    \item[] Guidelines:
    \begin{itemize}
        \item The answer NA means that the paper poses no such risks.
        \item Released models that have a high risk for misuse or dual-use should be released with necessary safeguards to allow for controlled use of the model, for example by requiring that users adhere to usage guidelines or restrictions to access the model or implementing safety filters. 
        \item Datasets that have been scraped from the Internet could pose safety risks. The authors should describe how they avoided releasing unsafe images.
        \item We recognize that providing effective safeguards is challenging, and many papers do not require this, but we encourage authors to take this into account and make a best faith effort.
    \end{itemize}

\item {\bf Licenses for existing assets}
    \item[] Question: Are the creators or original owners of assets (e.g., code, data, models), used in the paper, properly credited and are the license and terms of use explicitly mentioned and properly respected?
    \item[] Answer: \answerNA{} 
    \item[] Justification: The paper does not use existing assets.
    \item[] Guidelines:
    \begin{itemize}
        \item The answer NA means that the paper does not use existing assets.
        \item The authors should cite the original paper that produced the code package or dataset.
        \item The authors should state which version of the asset is used and, if possible, include a URL.
        \item The name of the license (e.g., CC-BY 4.0) should be included for each asset.
        \item For scraped data from a particular source (e.g., website), the copyright and terms of service of that source should be provided.
        \item If assets are released, the license, copyright information, and terms of use in the package should be provided. For popular datasets, \url{paperswithcode.com/datasets} has curated licenses for some datasets. Their licensing guide can help determine the license of a dataset.
        \item For existing datasets that are re-packaged, both the original license and the license of the derived asset (if it has changed) should be provided.
        \item If this information is not available online, the authors are encouraged to reach out to the asset's creators.
    \end{itemize}

\item {\bf New assets}
    \item[] Question: Are new assets introduced in the paper well documented and is the documentation provided alongside the assets?
    \item[] Answer: \answerNA{} 
    \item[] Justification: The paper does not release new assets.
    \item[] Guidelines:
    \begin{itemize}
        \item The answer NA means that the paper does not release new assets.
        \item Researchers should communicate the details of the dataset/code/model as part of their submissions via structured templates. This includes details about training, license, limitations, etc. 
        \item The paper should discuss whether and how consent was obtained from people whose asset is used.
        \item At submission time, remember to anonymize your assets (if applicable). You can either create an anonymized URL or include an anonymized zip file.
    \end{itemize}

\item {\bf Crowdsourcing and research with human subjects}
    \item[] Question: For crowdsourcing experiments and research with human subjects, does the paper include the full text of instructions given to participants and screenshots, if applicable, as well as details about compensation (if any)? 
    \item[] Answer: \answerNA{} 
    \item[] Justification: This paper does not involve crowdsourcing nor research with human subjects.
    \item[] Guidelines:
    \begin{itemize}
        \item The answer NA means that the paper does not involve crowdsourcing nor research with human subjects.
        \item Including this information in the supplemental material is fine, but if the main contribution of the paper involves human subjects, then as much detail as possible should be included in the main paper. 
        \item According to the NeurIPS Code of Ethics, workers involved in data collection, curation, or other labor should be paid at least the minimum wage in the country of the data collector. 
    \end{itemize}

\item {\bf Institutional review board (IRB) approvals or equivalent for research with human subjects}
    \item[] Question: Does the paper describe potential risks incurred by study participants, whether such risks were disclosed to the subjects, and whether Institutional Review Board (IRB) approvals (or an equivalent approval/review based on the requirements of your country or institution) were obtained?
    \item[] Answer: \answerNA{} 
    \item[] Justification: This paper does not involve crowdsourcing nor research with human subjects.
    \item[] Guidelines:
    \begin{itemize}
        \item The answer NA means that the paper does not involve crowdsourcing nor research with human subjects.
        \item Depending on the country in which research is conducted, IRB approval (or equivalent) may be required for any human subjects research. If you obtained IRB approval, you should clearly state this in the paper. 
        \item We recognize that the procedures for this may vary significantly between institutions and locations, and we expect authors to adhere to the NeurIPS Code of Ethics and the guidelines for their institution. 
        \item For initial submissions, do not include any information that would break anonymity (if applicable), such as the institution conducting the review.
    \end{itemize}

\item {\bf Declaration of LLM usage}
    \item[] Question: Does the paper describe the usage of LLMs if it is an important, original, or non-standard component of the core methods in this research? Note that if the LLM is used only for writing, editing, or formatting purposes and does not impact the core methodology, scientific rigorousness, or originality of the research, declaration is not required.
    \item[] Answer: \answerNA{} 
    \item[] Justification: The core method development in this research does not involve LLMs as any important, original, or non-standard components.
    \item[] Guidelines:
    \begin{itemize}
        \item The answer NA means that the core method development in this research does not involve LLMs as any important, original, or non-standard components.
        \item Please refer to our LLM policy (\url{https://neurips.cc/Conferences/2025/LLM}) for what should or should not be described.
    \end{itemize}

\end{enumerate}
\end{document}